%% file: main.tex
\pdfoutput=1
\documentclass[11pt,onecolumn]{cleantechnicalreport}
\usepackage[T1]{fontenc}
\usepackage[authoryear,round]{natbib}
\usepackage{hybridfrontmatter}

\input{math_commands.tex}

\usepackage{amssymb}
\usepackage{booktabs}
\usepackage{tabularx}
\usepackage{graphicx}
\usepackage{xcolor}
\usepackage{colortbl}
\definecolor{ink}{HTML}{30323D}
\definecolor{proposed}{HTML}{00A087}
\definecolor{teacher}{HTML}{F39B7F}
\definecolor{student}{HTML}{8491B4}
\definecolor{bothcorrect}{HTML}{4DBBD5}
\definecolor{baseline}{HTML}{949AA5}
\usepackage{hyperref}
\usepackage{url}
\hypersetup{hidelinks}
\title{DuoOPD: Learning from Joint Teacher--Student Outcomes for Multi-Task On-Policy Distillation}
\input{authors.tex}
\input{duoopd-style.tex}

\begin{abstract}
On-policy distillation (OPD) trains a student on its own responses with token-level feedback from a stronger teacher, yet the teacher can fail on questions the student already answers correctly, and how often each model succeeds varies across tasks.
OPD ignores these outcomes and, on average, pushes down even the student's correct responses; gating feedback by student correctness fixes the direction but uses the teacher in the same way whether or not it succeeded.
We introduce DuoOPD, in which the student's outcome sets the direction of feedback and the joint teacher--student outcome decides how the teacher supports it: when only the teacher succeeds, its verified answer becomes context for scoring the student's failed response, and when only the student succeeds, a weight shared within the task reinforces the whole response.
A single rule covers all four outcome combinations without task-specific settings.
Across Qwen3 and Llama, DuoOPD outperforms all five baselines in mean macro accuracy, improving over OPD by 2.58 and 5.98 percentage points, and it also leads on two further task mixtures spanning scientific calculation, instruction following, and code generation.
Ablations show that outcome-based direction alone stays near the gated baseline, while the joint-outcome designs supply most of the gain.
\end{abstract}

\begin{document}

\maketitle

\section{Introduction}
\label{sec:introduction}

\begin{figure}[t]
    \centering
    \includegraphics[width=\linewidth]{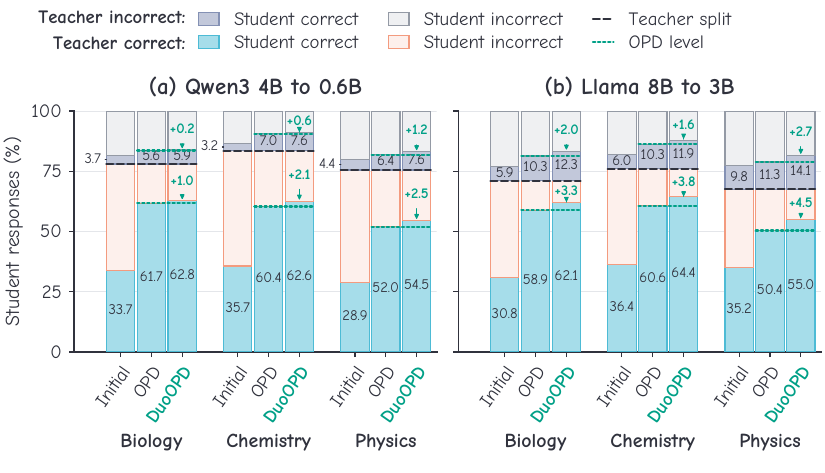}
    \caption{\small\textbf{DuoOPD improves student accuracy over OPD whether the teacher succeeds or fails:} (a) Qwen3; (b) Llama.
    Each bar splits student test responses by the fixed teacher response: below the dashed line are questions the teacher solves, above it those the teacher fails, so the line is identical for Initial, OPD, and DuoOPD. The lower segment of each part is student successes; on the DuoOPD bar, each green arrow gives the rise of the student-success segment below it above OPD's level (green dotted line, percentage points), positive in every domain for both parts. Each domain uses 200 shared questions with eight student responses per question; trained bars average three runs.}
    \label{fig:outcome-changes}
\end{figure}

On-policy distillation (OPD) trains a student on its own responses with token-level feedback from a stronger teacher \citep{lin2020imitkd,gu2024minillm,agarwal2024gkd}. When one teacher distills several tasks into one student, this feedback is not uniformly reliable: the teacher solves many questions the student misses, yet fails on others the student already answers correctly. Before distillation, our Qwen3 and Llama pairs show all four correctness combinations in task-dependent proportions (Initial bars in Figure~\ref{fig:outcome-changes}); the Llama student, for example, succeeds where the teacher fails on 5.9\% of biology responses but 9.8\% of physics responses. Across more diverse tasks the spread widens: both models fail on 1\% of chemistry-understanding training responses but 34\% of physics-calculation ones. Multi-task distillation therefore needs one rule that learns from teacher successes while reinforcing student successes when the teacher fails.

Existing methods leave this rule open. Multi-task methods decide which teacher supervises each task \citep{ma2026mopd,ma2026promptsd} or how strongly each task is trained \citep{ramesh2026mtgrpo}, not how feedback on a response depends on which model succeeded. OPD itself ignores correctness, so it can suppress correct student responses or reinforce failures; student verification can constrain feedback direction \citep{lin2026opdvr}, and teacher reliability probes can route prompts between OPD and reinforcement learning \citep{zhang2026tgopd}. Direction alone is not enough: student prefixes can steer even a capable teacher toward an incorrect solution \citep{zhu2026manyfaces}, and teacher-dependent positive weights can leave parts of successful responses weakly reinforced. The teacher's scoring context and allocation of feedback must also depend on who succeeded.

\begin{figure}[t]
    \centering
    \includegraphics[width=\linewidth]{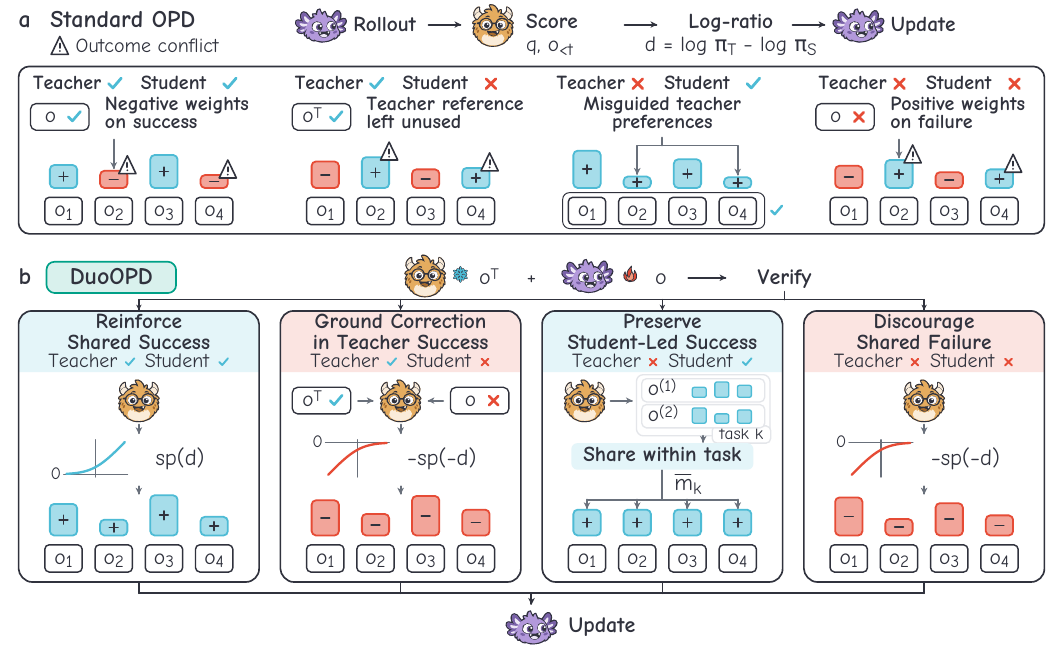}
    \caption{\small\textbf{Student outcomes set direction; joint outcomes select teacher use.}
    \textbf{a,} In OPD, triangles mark feedback opposing student outcomes, and a failing teacher's preferences weakly reinforce parts of correct responses.
    \textbf{b,} DuoOPD selects teacher use by joint outcome (Eq.\eqref{eq:feedback}); only the teacher reads $o^T$, $d$ is the teacher--student log-ratio, $\operatorname{sp}$ is Softplus, and shared weights follow Eq.\eqref{eq:online-mean}. Snowflake/flame mark frozen/trainable models; weight heights are schematic.}
    \label{fig:outcome-conditioning}
\end{figure}

Our key idea is to let student outcomes set the learning goal and joint outcomes determine how the teacher supports it. DuoOPD implements this principle as one four-outcome rule shared across tasks (Figure~\ref{fig:outcome-conditioning}): when only the teacher succeeds, a teacher reference guides correction; when only the student succeeds, weight sharing reinforces its complete response; when both succeed or both fail, teacher preferences set the strength of reinforcement or suppression. When the models disagree, support comes from the model that succeeded. Because the rule conditions on each response's outcome rather than its task, it needs no task-specific settings.

We evaluate three task mixtures of increasing heterogeneity, up to physics answers, instruction following, and code generation with their own verifiers. In every domain for both model families, DuoOPD raises student accuracy above OPD, both on questions the teacher solves and on questions it fails (Figure~\ref{fig:outcome-changes}). Our contributions are:
\begin{itemize}
    \item A unified approach to multi-task on-policy distillation from a single teacher into a single student: one four-outcome rule, free of task-specific settings, in which student verification sets feedback direction and joint outcomes select teacher context and weight allocation.
    \item Consistent gains across model families and task mixtures. DuoOPD leads all five baselines for Qwen3 and Llama, improving mean macro accuracy over OPD by 2.58 and 5.98 percentage points, and also leads on two additional mixtures spanning scientific understanding and calculation, instruction following, and code generation.
    \item Evidence for where the gains come from: every outcome rule contributes, and while outcome-based direction is necessary because outcome-agnostic OPD pushes down correct student responses, it alone stays near OPDVR; teacher references and within-task weight sharing supply most of the gain over OPD.
\end{itemize}

\section{Problem Setup and Background}
\label{sec:setup}

\paragraph{Multi-task setting.}
For each task $k\in\{1,\ldots,K\}$, $\mathcal D_k$ contains questions $q$ and ground-truth answers $a^\star$. We jointly train a student $\pi_\theta$ on a fixed task mixture with a frozen teacher $\pi_T$.

\paragraph{On-policy distillation.}
On-policy distillation obtains teacher feedback on student-generated prefixes. GKD matches full next-token distributions under a generalized divergence \citep{agarwal2024gkd}, and MiniLLM optimizes sequence-level reverse KL with policy gradients \citep{gu2024minillm}. We use the per-token sampled form common in recent practice \citep{lu2025onpolicydistillation}: at each update, the student with parameters $\theta_{\mathrm{old}}$ generates $o\sim\pi_{\theta_{\mathrm{old}}}(\cdot\mid q)$. Teacher and student score each sampled token under its prefix, giving the feedback \citep{zhu2026manyfaces}:
\begin{equation}
    d_{0,t}=\log\pi_T(o_t\mid q,o_{<t})
          -\log\pi_{\theta_{\mathrm{old}}}(o_t\mid q,o_{<t}).
    \label{eq:ratio}
\end{equation}
Subscript $0$ denotes scoring under $q$ and $o_{<t}$. At a fixed prefix, weighting the negative student log-probability gradient by this log-ratio gives a sampled-token estimator of the reverse-KL gradient. Each sampled token thus receives one scalar weight, which DuoOPD reshapes according to joint outcomes. This formulation is our OPD baseline.

For a batch of $B$ responses, index quantities by response $i$ and let $T_i$ count its valid tokens. The sampled OPD gradient averages within responses and then equally across them:
\begin{equation}
    \widehat g_{\mathrm{OPD}}(\theta)=
    -\frac{1}{B}\sum_{i=1}^{B}\frac{1}{T_i}
      \sum_{t=1}^{T_i}d_{0,i,t}\,
      \nabla_\theta\log\pi_\theta(o_{i,t}\mid q_i,o_{i,<t}).
    \label{eq:opd-gradient}
\end{equation}
Responses and log-ratio weights are computed before the update using $\theta_{\mathrm{old}}$ and held fixed; gradients flow only through the student log probabilities. Under gradient descent, positive weights reinforce sampled tokens and negative weights suppress them.

\paragraph{Outcome verification and gated feedback.}
A task-specific verifier gives $r_S=V_k(o,a^\star)\in\{0,1\}$, where 1 denotes correctness. OPDVR \citep{lin2026opdvr} masks feedback opposing this outcome and retains aligned weights, replacing $d_{0,i,t}$ in Eq.\eqref{eq:opd-gradient} with $A_{\mathrm{gate}}(d_{0,i,t},r_{S,i})$:
\begin{equation}
    A_{\mathrm{gate}}(d,r_S)=
    \begin{cases}
        \max\{d,0\}, & r_S=1,\\
        \min\{d,0\}, & r_S=0
    \end{cases}.
    \label{eq:opdvr-gate}
\end{equation}
We also verify an independently generated teacher response $o^T\sim\pi_T(\cdot\mid q)$, obtaining $r_T=V_k(o^T,a^\star)$. The joint outcome $(r_T,r_S)$ thus records whether a verified teacher reference is available and whether the student succeeds. Its distribution $\rho_k(a,b)=\Pr(r_T=a,r_S=b\mid k)$ differs across tasks and shifts as the student learns.

\section{DuoOPD: Distillation Guided by Joint Outcomes}
\label{sec:method}

Before training, we cache and verify one independent teacher response per training question; these references and outcomes stay fixed while the student generates fresh responses at each update. DuoOPD replaces each log-ratio weight in Eq.\eqref{eq:opd-gradient} with a weight $A_{i,t}$ determined by the joint outcome of its response (Figure~\ref{fig:outcome-conditioning}).

\subsection{Principle: Direction from the Student, Support from the Pair}
\label{sec:principle}

DuoOPD separates two decisions for each student response: whether to reinforce or suppress it, and how to distribute that feedback over its tokens. The student outcome makes the first decision and the joint outcome makes the second:
\begin{equation}
    \begin{aligned}
        A_{i,t}&=\sigma_i\,m_{i,t},\qquad \sigma_i=2r_{S,i}-1,\\
        m_{i,t}&=
        \begin{cases}
            \operatorname{sp}(\sigma_i\,d_{0,i,t}), & r_{T,i}=r_{S,i},\\
            \operatorname{sp}(-d_{T,i,t}), & r_{T,i}=1,\ r_{S,i}=0,\\
            \bar m_{k(i)}, & r_{T,i}=0,\ r_{S,i}=1.
        \end{cases}
    \end{aligned}
    \label{eq:feedback}
\end{equation}
Here $\operatorname{sp}$ is Softplus (Section~\ref{sec:direction}), $d_{T}$ is the teacher--student log-ratio with the teacher's verified answer in its context (Eq.\eqref{eq:reference-ratio}), and $\bar m_k$ is a positive magnitude shared within task $k$ (Eq.\eqref{eq:online-mean}). Every magnitude is positive, so the sign of each weight is the student's verified outcome: all valid tokens of a success are reinforced and all tokens of a failure are suppressed, whatever the teacher prefers. The joint outcome decides where the magnitude comes from. When the two models agree, teacher preferences in the original context set the strength of reinforcement or suppression. When they disagree, support comes from the model that succeeded: a successful teacher lends its verified answer as context for scoring the student's failed response, and a successful student receives a shared weight that the failing teacher's token preferences cannot reshape.

Eq.\eqref{eq:feedback} is shared by all tasks and never estimates $\rho_k$: it applies to whatever mix of outcomes a task presents, and a task enters only through its verifier $V_k$ and the scope of weight sharing. Section~\ref{sec:direction} explains the direction rule and the two agreements; Section~\ref{sec:disagreements} details the disagreement designs.

\subsection{A Shared Feedback-Direction Rule}
\label{sec:direction}

The magnitude should follow the verified outcome, use the log-ratio to set strength, and stay nonzero for every valid token. Unit-temperature Softplus $\operatorname{sp}(x)=\log(1+\exp(x))$ meets these requirements, giving positive weights $A^+$ for student success and negative weights $A^-$ for failure:
\begin{equation}
    \begin{aligned}
        A^+(d)&=\operatorname{sp}(d)=\max\{d,0\}+\log(1+\exp(-|d|)),\\
        A^-(d)&=-\operatorname{sp}(-d)=\min\{d,0\}-\log(1+\exp(-|d|)).
    \end{aligned}
    \label{eq:softplus}
\end{equation}
The first terms are OPDVR's gated weights at the same $d$ (Eq.\eqref{eq:opdvr-gate}); the residual follows the verified direction, with magnitude $\log 2$ at $d=0$ and approaching zero as $|d|$ grows. Softplus thus preserves the gated contribution and supplies nonzero feedback where the gate returns zero. This matters in practice: the teacher--student log-ratio averages below zero even on verified successes, so gated weights on correct responses average only 0.08--0.09 on DuoOPD's training responses for both Qwen3 and Llama, versus 0.57--0.65 under Softplus (Appendix~\ref{app:gate-vs-softplus}). The gate would thus leave most successful responses almost unreinforced, and the within-task shared weight, an average of these magnitudes, would nearly vanish. Applying $A^\pm$ to $d_0$ for every outcome gives the ``w/o both'' variant in Table~\ref{tab:components}; retaining only the gated terms recovers OPDVR.

\paragraph{Reinforce Shared Success.}
When $r_{T,i}=r_{S,i}=1$, the learning goal is to reinforce the verified response. We use $A^+(d_{0,i,t})$ under the original teacher context: every valid token receives positive feedback, even one the teacher assigns lower probability than the student, and teacher preferences set the strength of this positive feedback.

\paragraph{Discourage Shared Failure.}
When $r_{T,i}=r_{S,i}=0$, no correct reference is available, and the learning goal is to discourage the failed response. We use $A^-(d_{0,i,t})$ under the original teacher context: tokens the teacher disfavors relative to the student receive stronger negative feedback, but verification fixes the direction, so these preferences adjust suppression strength without turning it into reinforcement. Appendix~\ref{app:agreement-cases} illustrates both agreements.

\subsection{Support When Teacher and Student Disagree}
\label{sec:disagreements}

\paragraph{Ground Correction in Teacher Success.}
When $r_{T,i}=1$ and $r_{S,i}=0$, the student response requires correction and a verified teacher reference is available. The student's erroneous prefix can bias teacher scoring despite the teacher's independent success, so we add the successful teacher reference $o_i^T$ to the teacher's context:
\begin{equation}
    d_{T,i,t}=\log\pi_T(o_{i,t}\mid q_i,o_i^T,o_{i,<t})
              -\log\pi_{\theta_{\mathrm{old}}}(o_{i,t}\mid q_i,o_{i,<t}).
    \label{eq:reference-ratio}
\end{equation}
$A^-(d_{T,i,t})$ keeps the negative direction set by the student's failure, while the reference-conditioned teacher distribution allocates correction across the unchanged student trajectory. As in OPSD and OPCD \citep{zhao2026opsd,ye2026opcd}, only the teacher sees the reference (Appendix~\ref{app:reference-prompt}).

\paragraph{Preserve Student-Led Success.}
When $r_{T,i}=0$ and $r_{S,i}=1$, the student provides the verified success. Positive Softplus ensures the right direction, but the failing teacher can still assign weaker feedback to particular tokens or responses. We therefore give all valid tokens of these responses one positive weight shared within each task. Let $k(i)$ denote the task of response $i$, and let $\mathcal I_k=\{i:k(i)=k,\ r_{T,i}=0,\ r_{S,i}=1\}$ index them in the complete rollout batch. For each nonempty $\mathcal I_k$,
\begin{equation}
    \bar m_k=\operatorname{sg}\left[
    \frac{1}{|\mathcal I_k|}\sum_{i\in\mathcal I_k}
    \frac{1}{T_i}\sum_{t=1}^{T_i}\operatorname{sp}(d_{0,i,t})
    \right].
    \label{eq:online-mean}
\end{equation}
Here $\operatorname{sg}$ stops gradients, and the mean is computed once per rollout batch. It reinforces each complete response with the same coefficient while preserving the task's mean reinforcement scale; we share within tasks because log-ratio magnitudes differ across tasks (Section~\ref{sec:sharing-control}).

\subsection{Student Update}

Replacing $d_{0,i,t}$ in Eq.\eqref{eq:opd-gradient} with $A_{i,t}$ from Eq.\eqref{eq:feedback} gives the gradient of the DuoOPD surrogate loss:
\begin{equation}
    \mathcal L_{\mathrm{DuoOPD}}(\theta)=
    -\frac{1}{B}\sum_{i=1}^{B}\frac{1}{T_i}
    \sum_{t=1}^{T_i}\operatorname{sg}[A_{i,t}]
    \log\pi_\theta(o_{i,t}\mid q_i,o_{i,<t}).
    \label{eq:duoopd-loss}
\end{equation}
Only student log probabilities receive gradients; teacher references affect $A_{i,t}$, not the student context. Appendix~\ref{app:optimization} gives the correspondence between this surrogate and the implemented update.

\section{Experiments}
\label{sec:experiments}

\subsection{Experimental Setup}

\paragraph{Data and Verification.}
Three task mixtures progressively widen task differences: (i) biology, chemistry, and physics knowledge differ in domain; (ii) materials knowledge, chemistry understanding, and physics calculation also differ in cognitive level; (iii) physics knowledge, instruction following, and code generation also differ in output format and verifier. The first two use fixed, disjoint training and test partitions built from SciKnowEval V2 \citep{feng2024sciknoweval}. The third reuses the physics partition and adds RLVR-IFeval \citep{lambert2024tulu3} for training and official IFEval for testing \citep{zhou2023instruction}, and MBPP \citep{austin2021program}. Verification is binary: answer matching, instruction checks, and unit tests. All methods share data and verifiers within each mixture (Appendix~\ref{app:data}).

\paragraph{Models and Budget.}
We compare Qwen3-4B $\rightarrow$ Qwen3-0.6B \citep{yang2025qwen3} and Llama-3.1-8B-Instruct $\rightarrow$ Llama-3.2-3B-Instruct \citep{grattafiori2024llama3} on the initial mixture; both additional mixtures use the Qwen3 pair. Within each mixture, all methods process the same fixed-order question stream for 60 student updates, with 48 questions per update (16 per task), one sampled student response per question, and a 1,024-token response limit. Smaller task pools cycle to fill this budget. Appendix~\ref{app:implementation} reports implementation settings.

\paragraph{Teacher Computation.}\label{sec:cost}
DuoOPD caches one teacher response per distinct training prompt. This one-time cache takes 30\% (Qwen) and 23\% (Llama) of an OPD training loop in GPU-time and is reusable across students, seeds, and hyperparameters. During training, only responses where the teacher alone succeeds (13--23\%) are scored with a reference. The scoring and verification stage, including communication, accounts for 5.9\% and 4.7\% of OPD's loop. DuoOPD's measured training-loop change is $-$0.7\% (Qwen) and +7.3\% (Llama), with student generation accounting for most of the Llama increase (Appendix~\ref{app:cost}).

\paragraph{Comparison Methods.}
OPD and OPDVR use the original-context feedback weights defined in Eq.\eqref{eq:ratio} and Eq.\eqref{eq:opdvr-gate}, respectively. Entropy-aware OPD (EOPD) adds an entropy-gated distillation term \citep{jin2026eopd}; ExOPD extrapolates teacher feedback relative to the initial student \citep{yang2026gopd}; and FiRe-OPD filters trajectories and reweights feedback using teacher confidence and student entropy \citep{fireopd2026}.

\paragraph{Evaluation.}
We evaluate each method's final checkpoint. For each test question, we sample eight responses and report \emph{avg@8}, the mean binary success rate over responses and questions within each task. IFEval counts a response as correct only if it satisfies every instruction constraint \citep{zhou2023instruction}. Macro is the equal-weight mean across the three tasks. All results average three training runs per method (standard deviations in Appendix~\ref{app:variability}); initial-student and teacher rows use the same protocol. Development trajectories for the scientific settings appear in Appendix~\ref{app:development-curves}.

\subsection{Main Results}

\subsubsection{Comparisons Across Model Families}
\label{sec:model-families}

DuoOPD leads all five baselines on biology, chemistry, and physics for both Qwen3 and Llama (Table~\ref{tab:main}), improving mean Macro over OPD by 2.58 and 5.98 percentage points. It also exceeds the strongest baselines, EOPD on Qwen3 and OPDVR on Llama, by 1.16 and 4.58 points. Physics shows the largest gains over OPD (3.73 and 7.27 points) and is the domain where only the teacher succeeds most often during training (24.8\% and 14.9\%), the outcome corrected with teacher references. Section~\ref{sec:outcome-analysis} examines where these gains come from.

\begin{table}[t]
\centering
\small
\color{ink}
\caption{\textbf{Results across model families.} Test avg@8 accuracy (\%); Llama models are Instruct variants. Bold and underlined scores mark the best and second-best distillation methods per column.}
\label{tab:main}
\setlength{\tabcolsep}{4.5pt}
\renewcommand{\arraystretch}{1.05}
\begin{tabularx}{\linewidth}{l*{8}{>{\centering\arraybackslash}X}}
\toprule
 & \multicolumn{4}{c}{\shortstack{\textbf{Qwen3}\\4B $\rightarrow$ 0.6B}} & \multicolumn{4}{c}{\shortstack{\textbf{Llama}\\3.1-8B $\rightarrow$ 3.2-3B}} \\
\cmidrule(lr){2-5}\cmidrule(lr){6-9}
\textbf{Method} & Bio. & Chem. & Phys. & \textbf{Macro} & Bio. & Chem. & Phys. & \textbf{Macro} \\
\midrule
Student (initial) & 37.38 & 38.88 & 33.38 & 36.54 & 36.63 & 42.44 & 44.94 & 41.33 \\
Teacher & 79.56 & 81.38 & 76.88 & 79.27 & 73.25 & 75.06 & 69.88 & 72.73 \\
\midrule
OPD & 67.38 & 67.42 & 58.38 & 64.39 & 69.19 & 70.90 & 61.75 & 67.28 \\
EOPD & \underline{68.06} & \underline{68.44} & \underline{60.92} & \underline{65.81} & 68.73 & 69.54 & 61.21 & 66.49 \\
OPDVR & 66.92 & 66.69 & 60.90 & 64.83 & \underline{69.75} & \underline{72.40} & \underline{63.90} & \underline{68.68} \\
ExOPD & 67.54 & 66.79 & 58.94 & 64.42 & 68.33 & 69.88 & 60.81 & 66.34 \\
FiRe-OPD & 67.00 & 67.23 & 59.50 & 64.58 & 69.50 & 71.81 & 61.60 & 67.64 \\
\midrule
\rowcolor{proposed!15}
\textbf{DuoOPD (ours)} & \textbf{68.60} & \textbf{70.19} & \textbf{62.10} & \textbf{66.97} & \textbf{74.42} & \textbf{76.33} & \textbf{69.02} & \textbf{73.26} \\
\bottomrule
\end{tabularx}
\end{table}

\subsubsection{Results on Additional Task Mixtures}
\label{sec:mixtures}

\begin{table}[t]
\centering
\small
\color{ink}
\caption{\textbf{Results on two additional task mixtures with Qwen3-4B $\rightarrow$ Qwen3-0.6B.} Left: materials knowledge, chemistry understanding, and physics calculation. Right: physics knowledge, IFEval (prompt-level strict), and MBPP. Protocol and marking follow Table~\ref{tab:main}.}
\label{tab:extensions}
\setlength{\tabcolsep}{4.5pt}
\renewcommand{\arraystretch}{1.05}
\begin{tabularx}{\linewidth}{l*{8}{>{\centering\arraybackslash}X}}
\toprule
 & \multicolumn{4}{c}{\shortstack{\textbf{Science}\\knowledge, understanding, calculation}} & \multicolumn{4}{c}{\shortstack{\textbf{Heterogeneous}\\answers, instructions, code}} \\
\cmidrule(lr){2-5}\cmidrule(lr){6-9}
\textbf{Method} & Mat. & Chem. & Phys. & \textbf{Macro} & Phys. & IFEval & MBPP & \textbf{Macro} \\
\midrule
Student (initial) & 32.63 & 56.44 & 24.38 & 37.81 & 32.25 & 54.99 & 20.69 & 35.98 \\
Teacher & 67.44 & 96.94 & 56.00 & 73.46 & 77.25 & 78.70 & 57.69 & 71.21 \\
\midrule
OPD & \textbf{55.50} & 91.71 & 39.77 & 62.33 & 59.29 & 50.95 & 31.41 & 47.22 \\
EOPD & \underline{55.00} & 92.75 & 36.94 & 61.56 & 59.56 & 50.25 & 30.88 & 46.90 \\
OPDVR & 53.98 & \underline{92.83} & 39.67 & 62.16 & 59.35 & 52.31 & \underline{31.83} & \underline{47.83} \\
ExOPD & 54.56 & \textbf{93.21} & \underline{40.08} & \underline{62.62} & \underline{59.81} & 49.72 & 30.95 & 46.83 \\
FiRe-OPD & 54.67 & 92.17 & 38.31 & 61.72 & 58.96 & \underline{52.89} & 30.74 & 47.53 \\
\midrule
\rowcolor{proposed!15}
\textbf{DuoOPD (ours)} & 54.65 & 92.46 & \textbf{43.33} & \textbf{63.48} & \textbf{61.75} & \textbf{53.30} & \textbf{32.69} & \textbf{49.24} \\
\bottomrule
\end{tabularx}
\end{table}

\paragraph{Materials knowledge, chemistry understanding, and physics calculation.}
Each additional mixture is trained separately with the same DuoOPD settings; tasks enter only through their verifiers. DuoOPD retains the highest mean Macro when joint training combines scientific knowledge, understanding, and calculation (Table~\ref{tab:extensions}, left). It reaches 63.48\%, 0.86 points above ExOPD, the strongest baseline. Its largest task-level advantage, 3.25 points over the strongest baseline, is in physics calculation, where only the teacher succeeds on 26.8\% of training responses and both models fail on 34\%, whereas both succeed on 89\% of chemistry-understanding responses.

\paragraph{Physics knowledge, instruction following, and code generation.}
DuoOPD also leads all five baselines in mean Macro when the tasks require different output formats and verification procedures (Table~\ref{tab:extensions}, right). It reaches 49.24\%, 1.41 points above OPDVR, and scores highest on every task. Instruction following shows how tasks interact under joint training: the initial student already satisfies most IFEval prompts, and DuoOPD best preserves this ability alongside physics and code (53.30\% versus 50.95\% for OPD), as IFEval has the mixture's highest share of responses where only the student succeeds (7.6\%), which weight sharing reinforces.

\paragraph{Gains on the hardest task.}
Multi-task methods such as MT-GRPO explicitly target worst-task performance (Section~\ref{sec:related}). Without such an objective, DuoOPD attains the highest accuracy on the hardest task of every setting: physics for Qwen3 and Llama (62.10 and 69.02, versus at most 60.92 and 63.90), physics calculation (43.33 versus 40.08), and MBPP (32.69 versus 31.83).

\subsection{Ablation Study}
\label{sec:ablation}

All ablations use three Qwen3 runs per condition. The four-rule and component ablations use biology, chemistry, and physics; the sharing-scope comparison covers all three task mixtures.

\paragraph{Contributions of the Four Guidance Rules.}\label{sec:attribution}
Replacing any one outcome's feedback with OPD's $d_0$ lowers mean Macro by 0.76--1.53 points (Table~\ref{tab:controls}), so every rule contributes, including the rule for responses where only the student succeeds, which cover only 6.9\% of training responses.

\begin{table}[t]
\centering
\small
\color{ink}
\caption{\textbf{Replacing one outcome's feedback with OPD.} Share is the outcome's fraction of DuoOPD training responses (\%). Accuracy in \%; in Tables~\ref{tab:controls}--\ref{tab:sharing}, bold and underlined values mark the best and second-best means. $\Delta$ is relative to DuoOPD.}
\label{tab:controls}
\setlength{\tabcolsep}{4pt}
\renewcommand{\arraystretch}{1.05}
\begin{tabularx}{\linewidth}{ll>{\centering\arraybackslash}Xrrrrrr}
\toprule
\textbf{Teacher} & \textbf{Student} & \textbf{Feedback replacement} & Share & Bio. & Chem. & Phys. & \textbf{Macro} & $\Delta$ \\
\midrule
\rowcolor{proposed!15}
\multicolumn{2}{l}{\textbf{DuoOPD}} & --- & 100 & \textbf{68.60} & \textbf{70.19} & \textbf{62.10} & \textbf{66.97} & 0.00 \\
\checkmark & \checkmark & $\operatorname{sp}(d_0)\to d_0$ & 56.9 & 67.40 & 68.63 & \underline{61.73} & 65.92 & $-$1.05 \\
\checkmark & $\times$ & $-\operatorname{sp}(-d_T)\to d_0$ & 21.5 & 68.21 & 67.81 & 60.29 & 65.44 & $-$1.53 \\
$\times$ & \checkmark & $\bar m_k\to d_0$ & 6.9 & \underline{68.38} & \underline{68.81} & 61.44 & \underline{66.21} & $-$0.76 \\
$\times$ & $\times$ & $-\operatorname{sp}(-d_0)\to d_0$ & 14.7 & 68.27 & 68.79 & 60.67 & 65.91 & $-$1.06 \\
\bottomrule
\end{tabularx}
\end{table}

\paragraph{Teacher References and Weight Sharing.}\label{sec:reference-control}
The two disagreement rules carry DuoOPD's distinctive designs (Table~\ref{tab:components}). Without the teacher reference, responses where only the teacher succeeds are scored in the original context, and mean Macro drops by 1.49 points; without weight sharing, responses where only the student succeeds keep token-level positive weights, and it drops by 0.71. Removing both lowers it by 2.41 points, to 64.56\%, close to OPDVR (64.83\%), which also sets feedback direction by student correctness. Outcome-based direction alone therefore explains little of DuoOPD's 2.58-point gain over OPD; the two joint-outcome designs supply the rest, with complementary contributions. Appendix~\ref{sec:cases} shows both on fixed student responses.

\begin{table}[t]
\centering
\small
\color{ink}
\begin{minipage}[t]{0.57\linewidth}
\centering
\caption{\textbf{Removing the two disagreement designs.} Feedback used for each disagreement outcome; the agreement rules are unchanged.}
\label{tab:components}
\footnotesize
\setlength{\tabcolsep}{3pt}
\renewcommand{\arraystretch}{1.15}
\begin{tabularx}{\linewidth}{X>{\centering\arraybackslash}p{1.55cm}>{\centering\arraybackslash}p{1.2cm}rr}
\toprule
\textbf{Variant} & \textbf{T}\,\checkmark\,\textbf{S}\,$\times$ & \textbf{T}\,$\times$\,\textbf{S}\,\checkmark & \textbf{Macro} & $\Delta$ \\
\midrule
\rowcolor{proposed!15}
\textbf{DuoOPD} & $-\operatorname{sp}(-d_T)$ & $\bar m_k$ & \textbf{66.97} & 0.00 \\
w/o sharing & $-\operatorname{sp}(-d_T)$ & $\operatorname{sp}(d_0)$ & \underline{66.26} & $-$0.71 \\
w/o reference & $-\operatorname{sp}(-d_0)$ & $\bar m_k$ & 65.48 & $-$1.49 \\
w/o both & $-\operatorname{sp}(-d_0)$ & $\operatorname{sp}(d_0)$ & 64.56 & $-$2.41 \\
\bottomrule
\end{tabularx}
\end{minipage}\hfill
\begin{minipage}[t]{0.40\linewidth}
\centering
\caption{\textbf{Sharing scope.} Macro (\%) with the student-success weight shared across or within tasks.}
\label{tab:sharing}
\footnotesize
\setlength{\tabcolsep}{3pt}
\renewcommand{\arraystretch}{1.15}
\begin{tabularx}{\linewidth}{X*{3}{>{\centering\arraybackslash}p{0.85cm}}}
\toprule
\textbf{Shared} & B/C/P & Sci. & Het. \\
\midrule
Across tasks, $\bar m$ & \underline{66.74} & \underline{63.22} & \underline{48.89} \\
\rowcolor{proposed!15}
\textbf{Within task} & \textbf{66.97} & \textbf{63.48} & \textbf{49.24} \\
\bottomrule
\end{tabularx}
\end{minipage}
\end{table}

\paragraph{Sharing Within Each Task.}\label{sec:sharing-control}
Weight sharing is where the multi-task setting enters the rule. Teacher log-ratios differ in scale across tasks: per-task shared weights average 0.56--0.57 on biology, chemistry, and physics, but 0.56--0.64 and 0.57--0.64 in the two additional mixtures, where physics calculation and code generation receive the largest weights. One weight shared across all tasks would impose a single scale; sharing within each task keeps every task's own reinforcement scale while removing the failing teacher's preferences inside that task. Within-task sharing outperforms sharing across all tasks in all three mixtures (Table~\ref{tab:sharing}), with larger margins in the two mixtures whose task scales differ more (0.26 and 0.35 versus 0.22 points).

\subsection{Analysis by Joint Outcome}
\label{sec:outcome-analysis}

\paragraph{Gains on questions the teacher solves and on those it fails.}
Figure~\ref{fig:outcome-changes} splits DuoOPD's gain over OPD by teacher outcome. In every domain for both families, DuoOPD adds student successes both where the teacher succeeds (1.90 points on Qwen3, 3.86 on Llama) and where it fails (0.67 and 2.12): the student learns more of what the teacher knows and succeeds more often beyond it.

\paragraph{OPD pushes down correct responses.}
On DuoOPD's training responses, the original-context log-ratio that OPD would use as feedback is negative on average even when the student is correct: $-1.31$ on Qwen3 and $-0.22$ on Llama when both models succeed, and $-1.24$ and $-0.25$ when only the student succeeds. Correct responses make up 64--69\% of training responses, so OPD suppresses much of what the student already gets right. Setting direction by the student outcome, as OPDVR and DuoOPD do, prevents this, but direction alone stays near OPDVR's accuracy (Table~\ref{tab:components}); the gain comes from how DuoOPD uses the teacher once the direction is fixed.

\paragraph{Teacher references sharpen correction.}
Teacher references guide the most frequent disagreement, where only the teacher succeeds; removing them costs 1.49 points (Table~\ref{tab:components}). In Qwen3 runs trained with references, this outcome's log-ratio and feedback weight average $-1.80$ and $-2.35$, compared with $-1.35$ and $-1.91$ on the trajectories of runs trained without references. Appendix~\ref{sec:cases} holds a failed student response fixed and shows how adding the reference strengthens correction on its erroneous tokens.

\section{Related Work}
\label{sec:related}

\paragraph{Multi-task learning and distillation.}
LLM-based agents now support applications such as learner behavior simulation for intelligent education \citep{gao2025agent4edu,gao2026theater}, and a single deployed model is often expected to serve many such tasks. Multi-task tuning pools supervision across tasks \citep{sanh2022t0,chung2022flan}; distillation integrates task-specific ensembles \citep{liu2019mtkd}, domain RL specialists in MOPD \citep{ma2026mopd}, or soft-prompt teachers in PromptSD \citep{ma2026promptsd}, and MT-GRPO adapts task weights and sampling to improve worst-task performance \citep{ramesh2026mtgrpo}. These methods act at the task level; DuoOPD acts within each response of a fixed mixture, so the two compose: $r_T$ can verify a task-specific teacher, task weights can scale DuoOPD's per-task feedback, and $\rho_k$ offers a signal for task scheduling. Even without a worst-task objective, DuoOPD leads on the hardest task of every setting (Section~\ref{sec:mixtures}).

\paragraph{On-policy distillation and feedback calibration.}
On-policy distillation supervises student-generated trajectories through full-distribution divergences, sequence-level reverse KL, or per-token sampled feedback \citep{lin2020imitkd,agarwal2024gkd,gu2024minillm,lu2025onpolicydistillation}. EOPD uses entropy to adjust distillation, ExOPD extrapolates teacher feedback relative to the initial student, and FiRe-OPD filters trajectories and reweights tokens by teacher confidence and student entropy \citep{jin2026eopd,yang2026gopd,fireopd2026}. DuoOPD instead uses verified teacher--student outcomes to set the direction and allocation of feedback.

\paragraph{Verified outcomes and supervision routing.}
Verification can route prompts between OPD and GRPO \citep{zhang2026tgopd} or calibrate token targets \citep{qu2026spot}. OPDVR masks feedback that conflicts with student correctness \citep{lin2026opdvr}; DuoOPD's Softplus direction rule keeps the reinforcement on verified successes that the gate nearly removes (Section~\ref{sec:direction}). Direction alone stays near OPDVR (Section~\ref{sec:ablation}); DuoOPD's gain comes from also verifying the teacher, which selects the scoring context and weight sharing.

\paragraph{Privileged context and outcome-dependent teacher use.}
Privileged-context distillation gives teachers information unavailable to students \citep{lopezpaz2016privileged,zhao2026opsd,ye2026opcd}; in thinking models, such context can suppress exploration and self-correction \citep{kaur2026rethinkingopsd}. DuoOPD gives a verified teacher reference only when the teacher alone succeeds, and only the teacher reads it: the student's context is unchanged, and the reference only allocates negative feedback. H$^2$SD also separates student failures from successes, using hint-conditioned self-distillation and token-weighted reinforcement \citep{cai2026h2sd}; DuoOPD additionally verifies an independent frozen teacher, whose success supplies the reference and whose failure triggers weight sharing on correct student responses.

\section{Limitations and Future Work}
\label{sec:limitations}

DuoOPD relies on a binary verifier; open-ended tasks such as writing call for feedback rules that handle graded or uncertain judgments. Teacher computation (Section~\ref{sec:cost}) could shrink further through shared caches, compressed references, and selective reference use, and longer reasoning trajectories warrant study of exploration and self-correction. Each mixture is also trained and evaluated on the same tasks; whether joint-outcome guidance improves transfer to held-out tasks, and whether task-level outcome statistics $\rho_k$ can guide task scheduling, remain open.

\section{Conclusion}
\label{sec:conclusion}

DuoOPD lets the student's outcome set the direction of feedback and the joint outcome decide how the teacher supports it: teacher references correct failures the teacher can solve, and within-task weight sharing reinforces successes the teacher misses. Because the rule conditions on each response's outcome rather than its task, the same settings serve all three mixtures we test, which differ in domain, cognitive level, output format, and verifier. DuoOPD leads all five baselines across two model families and three task mixtures, including on the hardest task of every setting, and improves mean Macro over OPD by 2.58 points on Qwen3 and 5.98 on Llama. Our analyses show that setting direction by the student outcome is necessary, since outcome-agnostic OPD pushes down much of what the student already answers correctly, but not sufficient: in ablations, most of the gain comes from the two designs for disagreements.

\subsection*{Reproducibility statement}
Sections~\ref{sec:setup} and~\ref{sec:method} specify reference routing, token feedback, and loss aggregation. Appendices~\ref{app:data} and~\ref{app:implementation} provide data partitioning, prompts, model settings, optimization, and evaluation details. Section~\ref{sec:experiments} defines the evaluation protocol.
The DuoOPD repository (\url{https://github.com/YongYuanDeAo/DuoOPD}) includes code and setup instructions, processed data splits with the scripts that rebuild them from the source datasets, verifiers, cached teacher responses with generation settings, and five representative experiment configurations.

\bibliography{references}
\bibliographystyle{plainnat}

\clearpage
\appendix
\setcounter{figure}{0}
\renewcommand{\thefigure}{A\arabic{figure}}
\renewcommand{\theHfigure}{appendix.\arabic{figure}}
\section{Data and Evaluation Protocols}
\label{app:data}

\subsection{Biology, Chemistry, and Physics}
\label{app:science-data}

We use the public SciKnowEval V2 question collection (\texttt{hicai-zju/SciKnowEval}) to construct controlled multi-task distillation tasks. The official release provides a single \texttt{test} collection and no training or development split.\footnote{\href{https://huggingface.co/datasets/hicai-zju/SciKnowEval/blob/92ef969ad0a8bd6e195e0ac18af2c46e307e0cc2/README.md}{Pinned SciKnowEval V2 data card.}} We define our own fixed training, development, and test partitions and share them across all compared methods.

We retain the L1 biology and chemistry literature multiple-choice questions and four-option physics literature questions. We group duplicate questions across domains using Unicode NFKC normalization, case folding, and collapsed whitespace, and exclude groups with conflicting answers or domain assignments. Each domain contributes 960 training, 100 development, and 200 test questions, with partitions fixed before training.

A cross-partition audit of normalized question stems, excluding choices and the common instruction, finds no exact matches or pairs with character 5-gram Jaccard similarity at least 0.8.

The question and options are preserved, while the original instruction is replaced with:
\begin{quote}
\small
Explain the key reasoning briefly, then give only the final option letter (A, B, C, or D) in \texttt{\textbackslash boxed\{...\}}.
\end{quote}
The ground-truth answer and teacher reference are absent from student inputs. A response is scored as correct when its final boxed option matches the ground-truth letter. Missing or malformed boxes and invalid options receive zero credit.

\subsection{Materials, Chemistry, and Physics}
\label{app:science-extension}

The second mixture uses Material L1 knowledge, Chemistry L2 understanding, and Physics L3 calculation from SciKnowEval V2, with the same response instruction and answer-matching rule as Appendix~\ref{app:science-data}. Each task has 100 development and 200 test questions, shared across all six methods. The fixed 60-update training stream visits 960 distinct materials questions, 326 chemistry questions, and 474 physics questions, with 16 questions per task at each update. Smaller pools are cycled and reshuffled when exhausted. Other settings follow Table~\ref{tab:settings}.

\subsection{Physics, Instruction Following, and Code Generation}
\label{app:heterogeneous-extension}

\paragraph{Task sources and partitions.}
The physics task reuses the initial SciKnowEval V2 partition. Instruction-following training uses \texttt{allenai/RLVR-IFeval} from the T\"ulu~3 release \citep{lambert2024tulu3}, while evaluation uses the official IFEval prompts and checkers \citep{zhou2023instruction}.\footnote{Pinned sources: \href{https://huggingface.co/datasets/allenai/RLVR-IFeval/tree/47c03c73621c4aab2b824b7818681117d662770e}{RLVR-IFeval training data} and \href{https://github.com/google-research/google-research/tree/e1a63cd0666f970fc5df3753e17456c9f011d2a1/instruction_following_eval}{official IFEval evaluation}.} MBPP uses the original Google Research release with its official training, validation, and test assignments \citep{austin2021program}.\footnote{\href{https://github.com/google-research/google-research/blob/f82046ba5aabbbb427dbfd38a254d26bff08b533/mbpp/mbpp.jsonl}{Pinned MBPP data.}}

\paragraph{Instruction-following data and verification.}
Training uses eight constraint types: keyword inclusion, keyword frequency, bullet count, paragraph count, word count, JSON format, uppercase, and lowercase. We retain original user prompts, exclude choice and label-classification requests, and restrict keyword counts to 1--8, bullet and paragraph counts to 2--8, and word-count constraints to 10--150. The training pool contains 750 distinct prompts from 552 base questions; all variants of held-out base questions are excluded. Each update includes two prompts per constraint type, cycling smaller pools as needed.

Training responses are scored by whether they satisfy the specified constraint. We use the released RLVR checkers, with uppercase and lowercase checks requiring at least one cased character. Official IFEval evaluation uses prompt-level strict scoring: every constraint in the original prompt must be satisfied. Normalized prompt and base-question matching, including containment checks for base questions of at least 64 characters, finds no training--test overlap under these checks.

\paragraph{Code data and verification.}
We remove MBPP problems with duplicate descriptions within or across splits before model generation, leaving 371 training problems and 90 validation problems. All retained reference implementations pass the original tests. Prompts contain the task description, test setup, and first assertion to specify the interface, and request a Python code block. Verification executes generated code with the setup and all three original assertions in an isolated process; errors and timeouts count as failures.

\paragraph{Training and evaluation.}
Settings follow Table~\ref{tab:settings}. Physics visits 960 distinct training questions, while the instruction and code pools cycle as needed. The test set contains 200 physics questions, all 541 official IFEval prompts, and 496 MBPP problems after duplicate removal.

\section{Implementation and Computational Cost}
\label{app:implementation}

\subsection{Teacher Cache and Reference Prompt}
\label{app:reference-prompt}

For DuoOPD, we cache one independently generated teacher response per distinct training prompt, using temperature one, top-$p$ one, no top-$k$ restriction, seed 42, and a 1,024-token response limit. Responses are labeled using the task's answer-matching, instruction-checking, or code-testing rule (Appendix~\ref{app:data}). Cached responses and labels remain fixed during training.

When the cached teacher response passes the task check and the student's response fails it, the complete teacher response is supplied as a reference. Before applying the chat template, we append the following text to the teacher's last user message:
\begin{quote}
\small\ttfamily
Use the following verified reference to solve the question.\\
<reference>\\
\normalfont\itshape [Full teacher response that passes the task check]\\
\normalfont\ttfamily </reference>
\end{quote}
Other outcomes use original-context scoring; the teacher prompt limit is 5,632 tokens.

\paragraph{Token alignment.}
\label{app:alignment}
Each teacher--student pair shares a tokenizer and vocabulary. Both models score the same student response token IDs at aligned response positions; only the teacher context can include a reference. The loss includes response tokens and actual EOS tokens, excluding prompts, references, and padding.

\subsection{Optimization and Implementation}
\label{app:optimization}

\begin{table}[ht]
\centering
\small
\caption{Settings for the initial biology, chemistry, and physics comparison. Additional-mixture differences are specified in Appendices~\ref{app:science-extension} and~\ref{app:heterogeneous-extension}.}
\label{tab:settings}
\begin{tabularx}{\linewidth}{lX}
\toprule
Setting & Main comparison \\
\midrule
Teacher / initial student & Qwen3-4B / Qwen3-0.6B; Llama-3.1-8B-Instruct / Llama-3.2-3B-Instruct; teachers frozen \\
Optimizer & AdamW; $\beta=(0.9,0.95)$, weight decay $0.1$, $\epsilon=10^{-8}$ \\
Learning rate / schedule & $3\times10^{-6}$; cosine decay; 6 warmup steps \\
Updates / global batch & 60 / 48; 16 questions per domain \\
Student sampling & One response per question; temperature 1; top-$p=1$; unrestricted top-$k$ \\
Prompt / response limits & 4,096 / 1,024 tokens for the student \\
Evaluation & Checkpoint after update 60; chat template; temperature 1; no top-$p$ or top-$k$ truncation; seed 42; Qwen3 thinking disabled \\
Aggregation & Mean over valid tokens per response, then mean over responses \\
\bottomrule
\end{tabularx}
\end{table}

OPD and DuoOPD share the verl v0.7.0 scoring and update implementation \citep{sheng2024hybridflow}, with one rollout batch and one optimizer update per step; student rollouts and evaluation sampling use vLLM \citep{kwon2023vllm}. Feedback weights are computed in FP32 without centering and held fixed during each update, as specified in Section~\ref{sec:setup}. Additional entropy bonuses, reference-model KL penalties, and rollout correction are disabled. Task-level shared weights are computed over the complete batch before microbatch splitting; tasks with no response where only the student succeeds need no shared weight, and task sampling proportions are unchanged. Losses use the response averaging specified in Table~\ref{tab:settings}.

The actor uses verl's policy-ratio loss. With one minibatch and one epoch per rollout batch, its on-policy branch sets the denominator log probability to the detached current log probability. Writing $\ell=\log\pi_\theta(o_{i,t}\mid q_i,o_{i,<t})$, the ratio $u=\exp(\ell-\operatorname{sg}[\ell])$ has value one, so clipping is inactive, and $\nabla_\theta[-\operatorname{sg}[A_{i,t}]u]=-\operatorname{sg}[A_{i,t}]\nabla_\theta\ell$. The implemented loss therefore has the same gradient as Eq.\eqref{eq:duoopd-loss}, though their scalar values differ.

EOPD retains an additional teacher-entropy-gated KL loss on its top-16 teacher support, with entropy threshold $0.8$ and auxiliary coefficient one. Its actor microbatch is one because the upstream auxiliary loss averages valid tokens within a microbatch. This makes that term average within a response before gradient accumulation averages responses; other methods use actor microbatch four.

ExOPD uses reward extrapolation factor $\lambda=1.25$ and the frozen initial student as its reference model. FiRe-OPD uses a trajectory-filtering percentile of 20 and teacher-confidence and student-entropy coefficients $\alpha=\beta=1$; its filtering threshold and weight normalization are computed within each actor microbatch of four responses. Both retain upstream synchronous rollouts and token-level rollout importance sampling with threshold five.

\subsection{Teacher Computation and Runtime}
\label{app:cost}
On the initial biology, chemistry, and physics mixture, Table~\ref{tab:teacher-cost} reports DuoOPD's training and teacher-cache cost relative to OPD. Only responses where the teacher alone succeeds receive a reference, and the extra scoring input is measured against scoring the same student responses without references. Runtimes are mean training-loop times over the 60 updates, covering generation, scoring, and updates and excluding initialization, checkpoint saving, evaluation, and queue time. The scoring and verification stage also includes reference preparation and worker communication. Cache generation is timed on the same basis and converted to GPU-time. Student generation accounts for most of the Llama runtime increase, with DuoOPD responses averaging 177 tokens versus 152 for OPD.

\begin{table}[ht]
\centering
\small
\caption{DuoOPD training and teacher-cache cost relative to OPD. The first three rows give per-run ranges; rows with cache compare combined GPU-time against OPD's training-loop GPU-time, amortizing one cache over $R$ runs.}
\label{tab:teacher-cost}
\begin{tabular}{lrr}
\toprule
Quantity & Qwen & Llama \\
\midrule
Cache generated / student-sampled tokens per run & 1.01--1.10$\times$ & 0.76--0.90$\times$ \\
Responses scored with a reference per run & 20.7--22.6\% & 13.3--14.6\% \\
Extra teacher scoring input per run & 14.16--15.45\% & 6.21--6.39\% \\
Scoring/verification share of OPD training loop & 5.9\% & 4.7\% \\
Training-loop change, cache excluded & $-$0.7\% & +7.3\% \\
Cache generation / one OPD training loop & 29.7\% & 23.0\% \\
Training loop + full cache ($R=1$) & +29.0\% & +30.3\% \\
Training loop + cache shared across $R=3$ & +9.2\% & +15.0\% \\
\bottomrule
\end{tabular}
\end{table}

\section{Additional Results}
\label{app:additional-results}

\subsection{Variability Across Training Runs}
\label{app:variability}

Table~\ref{tab:variability} reports the Macro mean and sample standard deviation for Tables~\ref{tab:main} and~\ref{tab:extensions}.

\begin{table}[ht]
\centering
\small
\caption{Macro test avg@8 (\%), mean $\pm$ standard deviation.}
\label{tab:variability}
\begin{tabular}{lcccc}
\toprule
Method & Qwen3, B/C/P & Llama, B/C/P & Qwen3, science & Qwen3, heterogeneous \\
\midrule
OPD & 64.39 $\pm$ 0.45 & 67.28 $\pm$ 1.25 & 62.33 $\pm$ 0.39 & 47.22 $\pm$ 0.36 \\
EOPD & 65.81 $\pm$ 0.54 & 66.49 $\pm$ 0.52 & 61.56 $\pm$ 0.58 & 46.90 $\pm$ 0.36 \\
OPDVR & 64.83 $\pm$ 0.89 & 68.68 $\pm$ 1.37 & 62.16 $\pm$ 0.17 & 47.83 $\pm$ 0.35 \\
ExOPD & 64.42 $\pm$ 0.37 & 66.34 $\pm$ 0.58 & 62.62 $\pm$ 0.01 & 46.83 $\pm$ 0.08 \\
FiRe-OPD & 64.58 $\pm$ 0.39 & 67.64 $\pm$ 0.79 & 61.72 $\pm$ 0.22 & 47.53 $\pm$ 0.02 \\
DuoOPD & 66.97 $\pm$ 0.26 & 73.26 $\pm$ 1.04 & 63.48 $\pm$ 0.37 & 49.24 $\pm$ 0.64 \\
\bottomrule
\end{tabular}
\end{table}

\subsection{Development Trajectories}
\label{app:development-curves}

\begin{figure}[ht]
\centering
\includegraphics[width=0.90\linewidth]{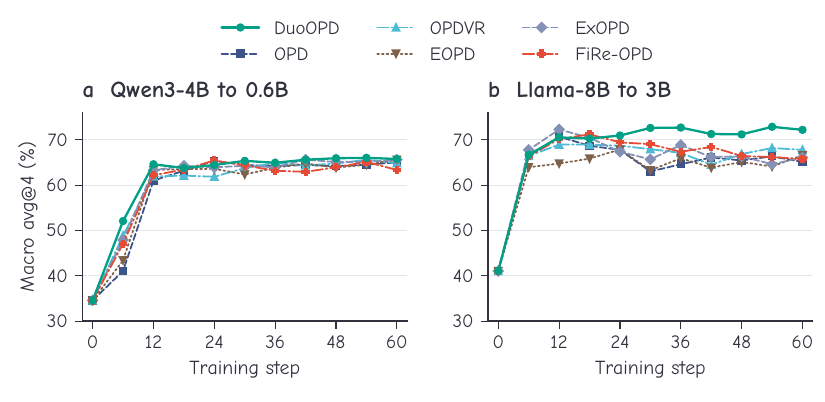}
\caption{Development Macro avg@4 on biology, chemistry, and physics for Qwen3-4B $\rightarrow$ Qwen3-0.6B (left) and Llama-3.1-8B-Instruct $\rightarrow$ Llama-3.2-3B-Instruct (right). Checkpoints every six updates from the shared initial student; both panels share one vertical scale.}
\label{fig:development}
\end{figure}

Figure~\ref{fig:development} tracks one training run per method on 100 development questions per task, with four sampled answers per question, every six updates from the shared initial student. In the displayed Llama runs, DuoOPD leads all baselines at every checkpoint from update 24 onward and stays above 70\%, while several baselines peak earlier and then decline. On Qwen3, DuoOPD has the highest development Macro at seven of the ten checkpoints, including the last three.

\section{How Joint Outcomes Change Token Feedback}
\label{app:cases}

We first compare feedback magnitudes under OPDVR's gate and DuoOPD's Softplus over all training responses, then illustrate on individual responses how teacher references and weight sharing allocate feedback and how student correctness determines its sign, each on a fixed student response.

\subsection{Gated versus Softplus Feedback Magnitudes}
\label{app:gate-vs-softplus}

Table~\ref{tab:gate-vs-softplus} compares, on the same training responses, the mean weight that OPDVR's gate and DuoOPD's Softplus would assign from the original-context log-ratio $d_0$. Statistics pool three DuoOPD runs per family; since $\max\{d,0\}=(d+|d|)/2$, the gated means follow from the recorded means of $d_0$ and $|d_0|$.

\begin{table}[ht]
\centering
\small
\caption{Mean token weight from $d_0$ under OPDVR's gate, $\sigma\max\{\sigma d_0,0\}$, and under Softplus, $\sigma\operatorname{sp}(\sigma d_0)$, where $\sigma=2r_S-1$. The gate keeps little positive feedback on verified successes.}
\label{tab:gate-vs-softplus}
\begin{tabular}{lcccccc}
\toprule
 & \multicolumn{2}{c}{Mean $d_0$} & \multicolumn{2}{c}{Gate} & \multicolumn{2}{c}{Softplus} \\
\cmidrule(lr){2-3}\cmidrule(lr){4-5}\cmidrule(lr){6-7}
Joint outcome & Qwen3 & Llama & Qwen3 & Llama & Qwen3 & Llama \\
\midrule
Both succeed & $-1.31$ & $-0.22$ & $+0.09$ & $+0.08$ & $+0.57$ & $+0.65$ \\
Only the student succeeds & $-1.24$ & $-0.25$ & $+0.09$ & $+0.09$ & $+0.57$ & $+0.64$ \\
Both fail & $-1.26$ & $-0.24$ & $-1.34$ & $-0.33$ & $-1.82$ & $-0.88$ \\
\bottomrule
\end{tabular}
\end{table}

\subsection{Reference-Guided Correction and Shared Reinforcement}
\label{sec:cases}

\begin{figure}[!ht]
    \centering
    \includegraphics[width=0.85\linewidth]{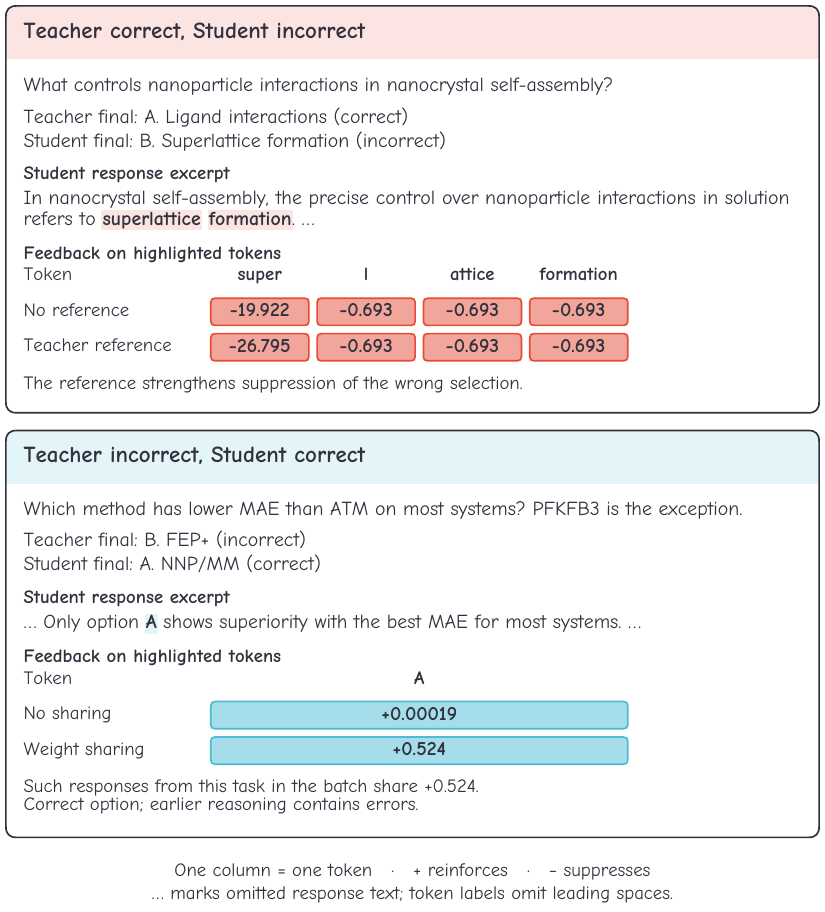}
    \caption{\textbf{Feedback allocation when teacher and student disagree.} Top: a teacher reference changes negative token weights. Bottom: weight sharing reinforces a correct option that otherwise receives almost no positive feedback. Numeric columns are tokens of the fixed response.}
    \label{fig:cases}
\end{figure}

\paragraph{Teacher correct, student incorrect: where to apply correction.}
In Figure~\ref{fig:cases} (top), the teacher selects ligand interactions, while the student selects superlattice formation. Adding the teacher reference strengthens negative feedback on ``super'' in the student's incorrect phrase. The student response and negative Softplus mapping stay fixed, isolating the reference's effect on correction.

\paragraph{Teacher incorrect, student correct: reinforce the correct selection.}
In Figure~\ref{fig:cases} (bottom), the student selects the correct NNP/MM answer despite the teacher's incorrect selection. Without sharing, the positive weight on ``A'' in ``Only option A'' is nearly zero; sharing gives it the task-level shared weight. This illustrates reinforcement of the correct selection: the student's earlier expansions of method names remain incorrect.

\clearpage
\subsection{Feedback Direction on Shared Success and Failure}
\label{app:agreement-cases}

\begin{figure}[ht]
\centering
\includegraphics[width=0.80\linewidth]{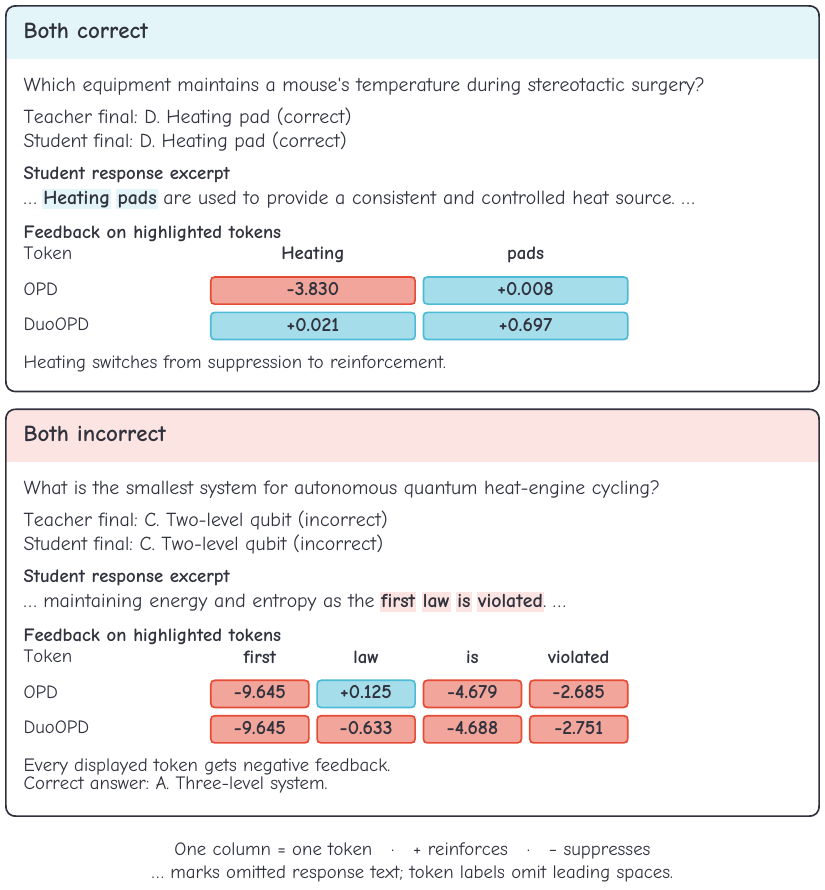}
\caption{\textbf{Feedback direction when teacher and student agree.} Positive Softplus reinforces the successful response; negative Softplus suppresses the failed response. Numeric columns are tokens of the fixed response.}
\label{fig:agreement-cases}
\end{figure}

\paragraph{Both correct: retain positive feedback.}
In Figure~\ref{fig:agreement-cases} (top), a token in the correct equipment name ``Heating pads'' receives OPD weight $-3.83$. Positive Softplus changes it to $+0.0215$, aligning feedback with the correct final answer.

\paragraph{Both incorrect: retain negative feedback.}
In Figure~\ref{fig:agreement-cases} (bottom), both models select a two-level qubit instead of the correct three-level system. The student's false assertion that the first law is violated receives mixed positive and negative OPD weights. Negative Softplus makes all displayed weights negative while preserving their ordering.

\paragraph{Case sources.}
Each case comes from the first rollout batch of Qwen3 training, when the student is still the initial model, and its token weights are computed with the initial teacher and student.

\end{document}

%% file: math_commands.tex
\usepackage{amsmath,amsfonts,bm}

\def\1{\bm{1}}

\DeclareMathAlphabet{\mathsfit}{\encodingdefault}{\sfdefault}{m}{sl}
\SetMathAlphabet{\mathsfit}{bold}{\encodingdefault}{\sfdefault}{bx}{n}

%% file: authors.tex
\newcommand{\DuoAuthorBlock}{%
  {\ReportAuthorFont
    Ao~Yu$^1$, Weibo~Gao$^2$, Heng~Zhou$^3$, Linan~Yue$^4$, Rui~Li$^1$,\\
    Suyi~Liu$^1$, Yu~Yan$^1$, Yizhong~Zhang$^1$, Qi~Liu$^{1\dagger}$\par}
  \vspace{5pt}
  {\ReportAffiliationFont
    $^1$University of Science and Technology of China\quad $^2$The Hong Kong Polytechnic University\par
    $^3$The University of Hong Kong\quad $^4$Southeast University\par}
  \vspace{6pt}
  {\normalfont\fontsize{8.5}{12}\selectfont
    \faEnvelope~\href{mailto:yuao@mail.ustc.edu.cn}{\textcolor{proposed}{\nolinkurl{yuao@mail.ustc.edu.cn}}}\quad
    $^{\dagger}$Corresponding author.\par
    \faGithub~\href{https://github.com/YongYuanDeAo/DuoOPD}{\textcolor{proposed}{\nolinkurl{https://github.com/YongYuanDeAo/DuoOPD}}}\par}
}
\hypersetup{pdfauthor={Ao Yu, Weibo Gao, Heng Zhou, Linan Yue, Rui Li, Suyi Liu, Yu Yan, Yizhong Zhang, Qi Liu}}

%% file: duoopd-style.tex
\colorlet{ReportAccent}{proposed}
\colorlet{ReportTheme}{proposed}
\colorlet{ReportAbstractBackground}{proposed!15!white}
\colorlet{ReportAbstractBorder}{proposed!50!white}
\renewcommand{\abscontent}{%
  \begin{tcolorbox}[
    colback=ReportAbstractBackground,colframe=ReportAbstractBorder,
    boxrule=0pt,arc=3pt,boxsep=0pt,
    left=12pt,right=12pt,top=12pt,bottom=12pt]
    {\centering\headingfont\color{ink}Abstract\par}
    \vspace{2mm}
    {\normalfont\fontsize{10}{13}\selectfont\theabstract\par}
  \end{tcolorbox}%
}
\hypersetup{hidelinks}
\titlespacing*{\paragraph}{0pt}{0pt}{1em}


\fancypagestyle{duoopdfirst}{%
  \fancyhf{}
  \fancyhead[L]{%
    \makebox[42pt][c]{\includegraphics[height=34pt,trim=5bp 6.33bp 7.67bp 6.33bp,clip]{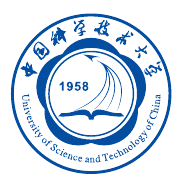}}\hspace{12pt}%
    \makebox[42pt][c]{\includegraphics[height=34pt,viewport=0 0 192 192,clip]{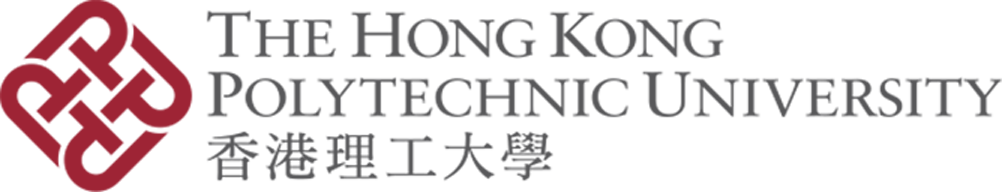}}\hspace{12pt}%
    \makebox[42pt][c]{\includegraphics[height=34pt,viewport=0 0 35.5 40,clip]{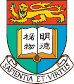}}\hspace{12pt}%
    \makebox[42pt][c]{\includegraphics[height=34pt]{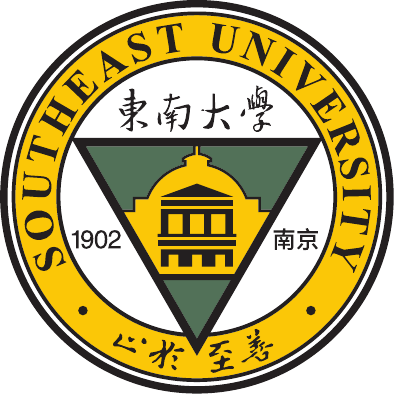}}}
  \fancyhead[R]{\footerfont\itshape 2026-09-27}
  
  \renewcommand{\headrule}{\color{ink}\DefaultHeadRule}
  
}
\makeatletter
\renewcommand{\maketitle}{%
  \begingroup
    \setlength{\parindent}{0pt}%
    \setlength{\parskip}{0pt}%
    {\raggedright\normalfont\bfseries\fontsize{19}{23}\selectfont\color{proposed}\@title\par}
    \vspace{10pt}
    {\raggedright\DuoAuthorBlock}
    \vspace{8pt}
    {\color{ink}\hrule height 0.6pt}
    \vspace{8pt}
    \abscontent
  \endgroup
  \thispagestyle{duoopdfirst}%
}
\makeatother